\def\year{2020}\relax
\documentclass[letterpaper]{article} 
\usepackage{aaai20}  
\usepackage{times}  
\usepackage{helvet} 
\usepackage{courier}  
\usepackage[hyphens]{url}  
\usepackage{graphicx} 
\usepackage{graphicx}  
\usepackage{bm}
\usepackage{amsfonts}       
\usepackage{amsmath}
\newtheorem{myDef}{Definition}

\usepackage{algorithm}
\usepackage{algorithmic}
\usepackage{booktabs}       
\usepackage{enumerate}
\usepackage{subfigure}

\title{Sparse-Dense Subspace Clustering}

\author{Shuai Yang, Wenqi Zhu, Yuesheng Zhu \\Institute of Big Data Technologies, \\ Shenzhen Graduate School, \\Peking University, China\\ \{ethanyang, wenqizhu, zhuys\}@pku.edu.cn}

\begin{document}

\maketitle

\begin{abstract}
  Subspace clustering refers to the problem of clustering high-dimensional data into a union of low-dimensional subspaces. Current subspace clustering approaches are usually based on a two-stage framework. In the first stage, an affinity matrix is generated from data. In the second one, spectral clustering is applied on the affinity matrix. However, the affinity matrix produced by two-stage methods cannot fully reveal the similarity between data points from the same subspace (intra-subspace similarity), resulting in inaccurate clustering. Besides, most approaches fail to solve large-scale clustering problems due to poor efficiency. In this paper, we first propose a new scalable sparse method called Iterative Maximum Correlation (IMC) to learn the affinity matrix from data. Then we develop Piecewise Correlation Estimation (PCE) to densify the intra-subspace similarity produced by IMC. Finally we extend our work into a Sparse-Dense Subspace Clustering (SDSC) framework with a dense stage to optimize the affinity matrix for two-stage methods. We show that IMC is efficient when clustering large-scale data, and PCE ensures better performance for IMC. We show the universality of our SDSC framework as well. Experiments on several data sets demonstrate the effectiveness of our approaches. Moreover, we are the first one to apply densification on affinity matrix before spectral clustering, and SDSC constitutes the first attempt to build a universal three-stage subspace clustering framework.
\end{abstract}

\section{Introduction}
High-dimensional data are ubiquitous in many applications of computer vision, e.g., face clustering \cite{1,2}, image representation and compression \cite{3}, and motion segmentation \cite{4,5}. A union of low-dimensional subspaces can approximate the original high-dimensional data for computational efficiency. The task of clustering high-dimensional data into corresponding low-dimensional subspaces is called subspace clustering.

Suppose that $X = \left[ \bm{x}_1,\dots,\bm{x}_N \right]\in \mathbb{R}^{D\times N}$ represents data set with $N$ data points in ambient dimension $D$, and data points lie in $n$ subspaces $\left\{S_i\right\}_{i=1}^{n}$ of dimensions $\left\{d_i\right\}_{i=1}^{n}$ ($d_i \ll \operatorname{min}\left\{D,N\right\}$). The task of subspace clustering is to partition data points into clusters $\left\{A_i\right\}_{i=1}^{n}$ so that data points within the same cluster $A_i$ lie in the same intrinsic subspace $S_i$. This problem has received great attention, and many algorithms including algebraic, iterative, statistical, and spectral clustering based approaches have been proposed (see \cite{6} for details). Among them, spectral clustering based methods have become extremely popular.

Current spectral clustering based methods solve the subspace clustering problem in two stages. In the first stage, an affinity matrix is learned to represent similarity between the data points. In the second stage, spectral clustering is applied on this affinity matrix. The differences of these methods lie in the first stage. These methods learn the affinity matrix based on the self-expressiveness model, which states that each data point in a union of subspaces can be expressed as a linear combination of other data points, i.e., $X = XC,$ where $X = [\bm{x}_1,\dots,\bm{x}_N]$ is the data matrix, $C = [\bm{c}_1,\dots,\bm{c}_N]\in \mathbb{R}^{N\times N}$ is the coefficients matrix. Once the $C$ is obtained, one can build an affinity matrix $W$ induced from $C$, e.g., $W = \left|C\right| + \left|C^T\right|$, and then obtain the segmentation of data by applying spectral clustering on $W$.

To find the coefficients matrix $C$, current methods solve the following optimization problem in the first stage:
\begin{equation}
\label{equ:1}
\min _{C}\|C\|_{C}, \text { s.t. } X=XC, \operatorname{diag}(C)= 0,
\end{equation}
where $\|\cdot\|_{C}$ denotes different norm regularization applied on $C$. For instance, in Sparse Subspace Clustering (SSC) \cite{7,8} and Structured Sparse Subspace Clustering (SSSC) \cite{9}, the $\ell_1$ is adopted as a convex surrogate over the $\ell_0$ norm to encourage the sparsity of $C$. Least Squares Regression (LSR) \cite{10} and Efficient Dense Subspace Clustering (EDSC) \cite{11} uses the $\ell_2$ norm regularization on $C$. Low Rank Representation (LRR) \cite{12}, Multiple Subspace Recovery (MSR) \cite{13} and Low-Rank Subspace Clustering (LRSC) \cite{14} use nuclear norm regularization on $C$. Low-Rank Sparse Subspace Clustering (LRSSC) \cite{15} uses a mixture of $\ell_1$ and nuclear norm regularization and Elastic Net Subspace Clustering (ENSC) \cite{16} uses a mixture of $\ell_1$ and $\ell_2$ regularization on $C$. In SSC by Orthogonal Matching Pursuit (OMP) \cite{17} and $\ell_0$-SSC \cite{18}, the $\ell_0$ norm is investigated. Block Diagonal Representation (BDR) \cite{19} uses block diagonal matrix induced regularizer to directly pursue the block diagonal matrix.

While the above approaches have been incredibly successful in many applications, we have observed some disadvantages. Firstly, The time cost of some approaches is too high to solve large-scale clustering problems, and the trade off between accuracy and efficiency is not the best. For example, SSC suffers from low efficiency and accuracy. SSSC improves the accuracy with the cost of extremely high computational time. BDR is much faster, but the accuracy is sometimes much lower than SSC.
Besides, 
for the sake of clustering, we expect the intra-subspace similarity to be as dense as possible, but computing a dense affinity matrix from data is extremely expensive when data lie in a high-dimensional space. More importantly, the disadvantage of the two-stage framework is the rough combination of computing affinity matrix and spectral clustering. Because the affinity matrix $W$ can not fully represent the relationship of data points, directly applying spectral clustering on this affinity matrix may result in poor accuracy. Motivated by these observations, we raise several interesting questions:
\begin{itemize}
\item Is it possible to find a sparse method to gain a better trade off between accuracy and efficiency, so that it can be applied on large-scale subspace clustering tasks?
\item Can we compute an affinity matrix with denser intra-subspace similarity in a more efficient way?
\item Is there a universal framework for those popular two-stage subspace clustering approaches that bridges the gap between the similarity computation and spectral clustering?
\end{itemize}

We aim to address the above questions and in particular we make the following contributions:
\begin{itemize}
\item We propose Iterative Maximum Correlation (IMC) to obtain a sparse affinity matrix. We show its efficiency and scalability when clustering 100,000 points, while most methods have only been tested at most 10,000 points.
\item We propose Piecewise Correlation Estimation (PCE) to densify the intra-subspace similarity in the affinity matrix produced by IMC. It estimates the similarity between two data points via other points. It is more efficient than directly computing a dense affinity matrix from data.
\item We extend our work to be a universal Sparse-Dense Subspace Clustering (SDSC) framework. We propose a dense stage to optimize the affinity matrix before spectral clustering, and it's universal for current popular methods. The dense stage ensures the better performance of SDSC.
\end{itemize}

To the best of our knowledge, we are the first one to densify the affinity matrix. Our SDSC constitutes the first attempt to build a three-stage subspace clustering framework and find a universal dense method for the affinity matrix.

We conduct experiments on synthetic data, the Extended Yale B \cite{20} face data set, the USPS \cite{21} and MNIST \cite{30} handwritten digits data sets. We show the scalability of our sparse approach IMC, the effectiveness of our dense method PCE, and the universality of our framework SDSC.

\section{Iterative Maximum Correlation (IMC)}
Recall from (\ref{equ:1}) that each data point in a union of subspaces can be expressed as a linear combination of other data points, different subspace clustering approaches uses different regularization methods to compute the coefficients matrix $C$. For the purpose of data clustering, we expect the coefficients matrix to be \emph{subspace-preserving} \cite{28} i.e., $c_{ij} \neq 0$ only if data points $\bm{x}_{i}$ and $\bm{x}_{j}$ lie in the same subspace. For computational efficiency, we relax the optimization problem (\ref{equ:1}) as the following program:
\begin{equation}
\label{equ:2}
\underset{C}{\min}\left\|X-XC\right\|_{2} \text { s.t. }\left\|C\right\|_{0} \leq \Lambda, diag(C) = 0,
\end{equation}
where $\Lambda$ constrains the number of nonzero entries in $C$. It is shown in \cite{29} that (\ref{equ:2}) can be efficiently solved by using greedy algorithms. This motivates us to propose Iterative Maximum Correlation (IMC) (Algorithm~\ref{alg:1}) to compute the sparse coefficients matrix from the original data.

Generally, for each point $\bm{x}_{i}$, IMC greedily selects the point $\bm{x}_j$ that is most linearly correlated with the residual, then fill the coefficient $c_{ij}$ in $C$ (step 5), and finally updates the residual by removing its projection on the most correlated vectorized data point $\bm{x}_{j}$ (step 6) until the iteration number reaches a certain value.

\begin{algorithm}[b]
\caption{Iterative Maximum Correlation (IMC)}
\label{alg:1}
\hspace*{\algorithmicindent} \textbf{Input:} Data set $X=[\bm{x}_1,\dots,\bm{x}_N]\in\mathbb{R}^{D\times N}$, IMC iteration number $\Gamma$.
\begin{algorithmic}[1]
\STATE \textbf{Initialize} coefficients matrix $C$ as a $N\times N$ zero matrix, index of current data point $i = 1$.
\WHILE{$i \leq N$}
\STATE \textbf{Initialize} current iteration $\gamma = 0$, residual $\bm{\psi}_0 = \bm{x}_i$.
\WHILE {$\gamma < \Gamma$}
\STATE $c_{ij} = \max\left|\rho_{\bm{\psi}_\gamma\bm{x}_j}\right|$, where $j \neq i$, and $\rho_{\bm{\psi}_\gamma\bm{x}_j}$ is calculated by (\ref{equ:3}).
\STATE Update residual $\bm{\psi}_{\gamma+1} = \bm{\psi}_{\gamma} - (\bm{\psi}_\gamma\cdot\bm{x}_j)\bm{x}_j$.
\STATE $\gamma \leftarrow \gamma + 1$.
\ENDWHILE
\STATE $i \leftarrow i + 1$
\ENDWHILE
\end{algorithmic}
\hspace*{\algorithmicindent} \textbf{Output:} The coefficients matrix $C$.
\end{algorithm}

In particular, IMC adopts the Pearson correlation coefficient \cite{22} to find the most linearly correlated point. It is the covariance of the two variables divided by the product of their standard deviations. It has been proven to be effective to measure the linear correlation between variables. Given two vectorized data points $\bm{x}_i$ and $\bm{x}_j$ in $\Delta$ dimension, the Pearson correlation coefficient $\rho_{\bm{x}_i\bm{x}_j}$ is calculated as:
\begin{equation}
\label{equ:3}
\rho_{\bm{x}_i\bm{x}_j}=\frac{\sum_{\delta=1}^{\Delta}\left(x_{i\delta}-\overline{x_i}\right)\left(x_{j\delta}-\overline{x_j}\right)}{\sqrt{\sum_{\delta=1}^{\Delta}\left(x_{i\delta}-\overline{x_i}\right)^{2}}\sqrt{\sum_{i=1}^{\Delta}\left(x_{j\delta}-\overline{x_j}\right)^{2}}},
\end{equation}
where $x_{i\delta}$ and $x_{j\delta}$ are the entries in vectors $\bm{x}_i$ and $\bm{x}_j$, $\overline{x_i} = \frac{1}{\Delta} \sum_{\delta=1}^{\Delta} x_{i\delta}$, and analogously for $\overline{x_j}$. The value of $\rho_{\bm{x}_i\bm{x}_j}$ is in the range between $-1$ and $+1$. The larger the absolute value $\left|\rho_{\bm{x}_i\bm{x}_j}\right|$, the stronger the linear relationship. An absolute value of 1 indicates a perfect linear relationship.

We obtain coefficients directly from IMC. The Pearson correlation coefficients are used not only as a measure to select data points, but also as the value of entries in coefficients matrix $C$. Before the $\gamma+1$th iteration, the residual $\bm{\psi}_{\gamma+1}$ is updated as the difference between the current residual $\bm{\psi}_{\gamma}$ and its projection on the most linearly correlated vector $\bm{x}_j$. Since $\bm{\psi}$ is affected by each iteration, and the contribution of all selected points are removed, it reduces the risk of duplicate selection of the same point.

In most current popular subspace clustering methods, the affinity matrix is computed to be symmetric by:
\begin{equation}
\label{equ:4}
W=\left(\left|C\right|+\left|C^{\top}\right|\right),
\end{equation}
where $C^{\top}$ denotes the transpose of $C$. However, for some mutually selected data points such as $\bm{x}_i$ and $\bm{x}_j$, the coefficients $c_{ij}$ and $c_{ji}$ are nonzero entries. Calculating $w_{ij}$ by $\left|c_{ij}\right|+\left|c_{ji}\right|$ changes the similarity obtained by IMC. Thus we propose a new way to construct the affinity matrix, that is:
\begin{equation}
\label{equ:5}
W=\max \left(\left|C\right|, \left|C^{\top}\right|\right),
\end{equation}
which means that for each $w_{ij}$, the value is the larger one of $\left|c_{ij}\right|$ and $\left|c_{ji}\right|$. Then we obtain the affinity matrix $W$.
\section{Piecewise Correlation Estimation (PCE)}
Before applying spectral clustering on the affinity matrix $W$ produced by IMC, we would like to analyze the similarity in $W$. To better interpret the similarity between data points, we divide the value of similarity into four levels. The levels and the corresponding range of similarity are defined next.

\begin{myDef}
\label{def:1}
(\textbf{Piecewise correlation}).
\begin{table}[h]
  \centering
  \small
    \begin{tabular}{l|l}
    \toprule
    Extremely strong correlation & $(\theta_1, 1]$ \\
    Strong correlation & $(\theta_2, \theta_1]$ \\
    Medium correlation & $(\theta_3, \theta_2]$ \\
    Weak correlation & $[0, \theta_3]$ \\
    \bottomrule
    \end{tabular}%
  \label{tab:addlabel}%
\end{table}\\
\noindent where $\theta_1$, $\theta_2$, $\theta_3$ are thresholds, and the value is to be fixed by the following experiments on real-world data sets.
\end{myDef}

We assume that a pair of data points $\bm{x}_i$ and $\bm{x}_j$ with strong or extremely strong correlation i.e., $w_{ij} > \theta_2$  indicates that they belong to the same subspace. With analysis of the affinity matrix $W$, we observe some similarity of pairwise data points from the same subspace is much smaller than expected, and some is even zero. We detailedly define this phenomenon as ternary unstable relationship next.

\begin{myDef}
\label{def:2}
(\textbf{Ternary unstable relationship}).
Given any two data points $\bm{x}_i$, $\bm{x}_j$ in $X$ and the similarity between them $w_{ij}$. Consider the pairwise correlation defined in Definition (\ref{def:1}), for any intermediate data point $\bm{x}_k$ $(\bm{x}_k \in X\backslash\left\{\bm{x}_i,\bm{x}_j\right\})$, we say that the relationship of $\bm{x}_i$, $\bm{x}_j$, and $\bm{x}_k$ is ternary unstable if their similarity $w_{ij}$, $w_{ik}$, $w_{kj}$ satisfies one of the following conditions (TUR conditions):
\begin{enumerate}[1)]
 \item $w_{ik}, w_{kj} \in (\theta_1, 1], w_{ij} \notin (\theta_1, 1];$
 \item $\max(w_{ik}, w_{kj})\in(\theta_1, 1], \min(w_{ik}, w_{kj})\in(\theta_2, \theta_1], \\w_{ij}\notin(\theta_2,1]$
 \item $w_{ik}, w_{kj} \in (\theta_2, \theta_1], w_{ij}=0.$
\end{enumerate}
\end{myDef}

Current spectral clustering approaches adopt normalized cut \cite{23} to partition the data points into $n$ clusters $\left\{A_i\right\}_{i=1}^{n}$. The dissimilarity between cluster $A_i$ and other clusters $\overline{A_i}$ are defined as $cut$:
\begin{equation}
\label{equ:6}
cut(A_i, \overline{A_i})=\sum_{x_u \in A_i, x_v \notin A_i} w_{uv},
\end{equation}
while the intra-cluster similarity of $A_i$ is defined as $vol$:
\begin{equation}
\label{equ:7}
vol(A_i)=\sum_{x_u, x_t \in A_i} w_{ut}.
\end{equation}
To measure the disassociation of $n$ clusters obtained by normalized cut, $Ncut$ is defined as:
\begin{equation}
\label{equ:8}
Ncut\left(A_{1}, A_{2}, \ldots A_{n}\right)= \sum_{i=1}^{n} \frac{cut\left(A_{i}, \overline{A_{i}}\right)}{vol\left(A_{i}\right)},
\end{equation}
and the task of normalized cut is to find how to partition the data points into $n$ clusters to get a minimum value of $Ncut$. An ideal affinity matrix for this task should contain as dense intra-subspace similarity as possible, and contain as few inter-subspace similarity elements as possible.

Recall from the $W$ produced by IMC, the ternary unstable relationship limits the intra-subspace similarity, making it difficult to group the points from the same subspace into the same cluster. This motivates us to revise the ternary unstable relationship to gain higher intra-subspace similarity.

We propose Piecewise Correlation Estimation (PCE) (Algorithm \ref{alg:2}) to revise the ternary unstable relationship in $W$. The similarity is optimized after traversal of all intermediate points. Due to the restriction of TUR conditions, for data points $\bm{x}_i$, $\bm{x}_j$ having stronger correlation with the intermediate point $\bm{x}_k$, the similarity $w_{ij}$ is updated more close to $w_{ik}$ and $w_{kj}$; for those with medium and weak correlation with the intermediate point, the similarity is not changed.

\begin{algorithm}[h]
\caption{Piecewise Correlation Estimation (PCE)}
\label{alg:2}
\hspace*{\algorithmicindent} {\textbf{Input:} $\theta_1$, $\theta_2$, $W$, data set $X$.}
\begin{algorithmic}[1]
\FOR{each pair of data points $\bm{x}_i$ and $\bm{x}_j$}
    \FOR {each intermediate point $\bm{x}_k \in X\backslash\left\{\bm{x}_i,\bm{x}_j\right\}$}
    \STATE
$w^*_{ij}=\left\{
             \begin{array}{ll}
             \frac{1}{2}(w_{ik} + w_{kj}),& \textbf{if}\text{ TUR condition1)}\\
             \min(w_{ik}, w_{kj}),& \textbf{if}\text{ TUR condition2)}\\
             \frac{1}{2}\max(w_{ik}, w_{kj}),& \textbf{if}\text{ TUR condition3)} \\
             w_{ij},& \textbf{else}
             \end{array}
\right.
$
    \ENDFOR
\ENDFOR
\end{algorithmic}
\hspace*{\algorithmicindent} \textbf{Output:} A new affinity matrix $W^*\in\mathbb{R}^{N\times N}$.
\end{algorithm}

After PCE, we can get $w^*_{ij} \geq w_{ij}$. In particular, some zero entries in $W$ are updated as nonzero entries, which densifies the affinity matrix. The new affinity matrix $W^*$ obtains larger intra-subspace similarity for each subspace $S_i$, while the inter-subspace similarity is slightly changed due to the restriction of TUR conditions. Recall from the task of normalized cut, points in $S_i$ are more likely to be clustered into the same cluster $A_i$ after PCE. Because from the perspective of $A_i$, the intra-cluster similarity $vol(A_i)$ is much increased and the inter-cluster similarity $cut(A_i, \overline{A_i})$ is just slightly increased. We will verify this by experiments. 
\section{Sparse-Dense Subspace Clustering (SDSC): \\A Universal Framework}
Directly combining IMC and spectral clustering as a two-stage approach is practical to solve subspace clustering problems, but inserting a dense stage PCE before spectral clustering ensures a better affinity matrix. This three-stage approach follows a Sparse-Dense Subspace Clustering (SDSC) framework as concluded in Algorithm \ref{alg:3}.

However, we observe that many other popular methods (e.g. SSC, LSR, LRR, OMP, ENSC, BDR) follow the two-stage framework that directly combining affinity matrix generation with spectral clustering. The affinity matrix obtained in the first stage usually contain insufficient intra-subspace similarity, which is difficult for spectral method to partition data points from the perspective of graph theory. We need an affinity matrix with as dense intra-subspace similarity elements as possible. It's difficult to obtain such an affinity matrix by these two-stage methods, since the computation is extremely expensive. For instance, SSSC integrates the two stages of SSC into a learning framework, and re-weights the similarity in many iterations, but the time cost is extremely high. Motivated by our proposed PCE for IMC, we attempt to optimize the affinity matrix before spectral clustering for current two-stage approaches, and finally remould those methods to follow the proposed SDSC framework.
\begin{algorithm}[h]
\caption{Sparse-Dense Subspace Clustering (SDSC)}
\label{alg:3}
\hspace*{\algorithmicindent} \textbf{Input:} Data set $X$.
\begin{algorithmic}[1]
\STATE Compute a affinity matrix $W$ from data by different data representation methods.
\STATE Optimize similarity in $W$ to get $W^*$ by a dense method.
\STATE Apply spectral clustering on $W^*$.
\end{algorithmic}
\hspace*{\algorithmicindent} \textbf{Output:} Clustering results.
\end{algorithm}

In SDSC, we use a dense stage to optimize the affinity matrix before spectral clustering. Different from PCE specifically proposed for IMC, this dense method needs to be universal for as many approaches as possible. We attempt to analyze and optimize the affinity matrix by distances graph, and propose a universal dense stage (Algorithm \ref{alg:4}) for current two-stage subspace clustering methods.

\begin{myDef}
\textbf{(Simulated distances)}.
To measure distances between data points, a distances matrix $D\in \mathbb{R}^{N\times N}$ is generated from the affinity matrix $W\in\mathbb{R}^{N\times N}$, and elements in $D$ represent the simulated distances between data points.
\end{myDef}

\begin{algorithm}
\caption{A universal dense stage for SDSC}
\label{alg:4}
\hspace*{\algorithmicindent} \textbf{Input:} $X$, affinity matrix $W$.
\begin{algorithmic}[1]
\STATE Compute simulated distances $D\in \mathbb{R}^{N\times N}$ from $W$.
\FOR{each pair of data points $\bm{x}_i$ and $\bm{x}_j$}
    \FOR {each intermediate point $\bm{x}_k \in X\backslash\left\{\bm{x}_i,\bm{x}_j\right\}$}
            \STATE {$d^*_{ij} = \min(d_{ij}, d_{ik} + d_{kj})$}
    \ENDFOR
\ENDFOR
\STATE Compute new affinity matrix $W^*$ from $D^*$.
\end{algorithmic}
\hspace*{\algorithmicindent} \textbf{Output:} The optimized affinity matrix $W^*\in \mathbb{R}^{N\times N}$.
\end{algorithm}

We first transform the similarity into simulated distances, then minimize the distances of each pair of points via the intermediate points, and finally transform the distances back into similarity. Step 1 in Algorithm \ref{alg:4} is the transformation form similarity to distances, and step 7 is the inverse transformation. The value of similarity is in the range $[0,1]$, and the similarity grows inversely to the distances. For the ease of use, we propose several practical transformation functions in Table \ref{tbl:1} and we will test them in experiments.

\begin{table}[htbp]
  \centering
  \small
  \caption{Several transformation functions.}
  \label{tbl:1}
    \begin{tabular}{l|l|l}
    \toprule
     &Step 1: $w_{ij}$ to $d_{ij}$ & Step 9: $d^*_{ij}$ to $w^*_{ij}$ \\
    \hline
    1) & $d_{ij} = 1 - w_{ij}$ & $w^*_{ij} = 1 - d^*_{ij}$ \\
    2) & $d_{ij} = 1 - \ln(w_{ij})$ & $w^*_{ij} = \exp(1 - d^*_{ij})$ \\
    3) & $d_{ij} = \frac{1}{w_{ij}}$ & $w^*_{ij} = \frac{1}{d^*_{ij}}$ \\
    \bottomrule
    \end{tabular}%
\end{table}%

Step 4 in Algorithm \ref{alg:4} minimizes the distance between $\bm{x}_i$ and $\bm{x}_j$ via the intermediate point $\bm{x}_k$. In particular, if $w_{ij}$ is zero in $W$, $w_{ik}$ and $w_{kj}$ are nonzero entries, then $w^*_{ij}$ are updated as nonzero after the dense stage. This makes the new affinity matrix $W^*$ a denser matrix.

Specifically, if one only needs to fine-tune the nonzero elements in $W$ without changing the sparsity, restrictive conditions of $d_{ij}$ can be added before step 4. This will not be detailed here since it makes limited changes to the affinity matrix, and makes little improvements of performance.

Given an affinity matrix $W$ of $N$ data points lying in $n$ subspaces, the sum of all the intra-subspace similarity elements in $W$ is $\eta_1$, and the sum of inter-subspace similarity is $\eta_2$. We respectively denote $\overline{\eta_1}$ as the average value of intra-subspace similarity in $W$, and $\overline{\eta_2}$ as the average value of inter-subspace similarity. There are $kN$ nonzero entries in $W$, where $k$ is the number of nonzero similarity elements of each data point. Particularly, suppose the number of inter-subspace similarity elements is $\zeta$, then the number of intra-subspace similarity elements in $W$ is $kN-\zeta$. Recall from the normalized cut problem in (\ref{equ:8}), we can calculate $Ncut$ for perfectly clustering each data point into intrinsic subspaces:
\begin{equation}
\label{equ:9}
Ncut = \frac{\eta_2}{\eta_1} = \frac{\zeta\overline{\eta_2}}{(kN-\zeta)\overline{\eta_1}}.
\end{equation}

For each nonzero entry $w_{ij}$ in $W$, there are other $k-1$ nonzero entries that in the $i$th row or the $j$th column. After the dense stage, the value of each entry is updated via the intermediate points. The number of nonzero entries is increased to be approximately $Nk^2 - \tau$, where $\tau$ is the number of duplicate intra-subspace entries. The number of inter-subspace similarity in $W^*$ is $\zeta (k-1)$, and the number of intra-subspace similarity elements is $Nk^2 -\zeta (k-1) - \tau$. We can calculate $Ncut^*$ for perfectly clustering of $W^*$:
\begin{equation}
\label{equ:10}
Ncut^* = \frac{\eta^*_2}{\eta^*_1} = \frac{\zeta (k-1)\overline{\eta^*_2}}{(Nk^2 -\zeta (k-1) - \tau)\overline{\eta^*_1}},
\end{equation}
where $\eta^*_1$ and $\eta^*_2$ are the sum of intra-subspace and inter-subspace similarity in $W^*$, $\overline{\eta^*_1}$ and $\overline{\eta^*_2}$ are the average value of them. The ratio of $Ncut^*$ to $Ncut$ can be computed as:
\begin{equation}
\label{equ:11}
Q = \frac{Ncut^*}{Ncut} = \frac{Nk^2 -\zeta (k-1) - Nk}{Nk^2 -\zeta (k-1)-\tau}\frac{\overline{\eta^*_2}\overline{\eta_1}}{\overline{\eta^*_1}\overline{\eta_2}}.
\end{equation}
Since $N >> \zeta$, the ratio $Q$ can be approximated as:
\begin{equation}
\label{equ:12}
\widetilde{Q} = \frac{Nk^2-Nk}{Nk^2-\tau}\frac{\overline{\eta^*_2}\overline{\eta_1}}{\overline{\eta^*_1}\overline{\eta_2}}.
\end{equation}
Generally, the average value of each intra-subspace and inter-subspace similarity element is slightly changed after the dense stage. Thus the ratio can be approximated as:
\begin{equation}
\label{equ:13}
\widetilde{Q} = \frac{Nk^2-Nk}{Nk^2 - \tau}.
\end{equation}
Actually $\tau$ is usually zero in real experiments, partly due to the high sparsity of the original affinity matrix $W$. The ratio $\widetilde{Q}$ is smaller than 1, i.e., $Ncut^* < Ncut$, indicating that the densified affinity matrix $W^*$ makes the points from the same subspace more likely to be clustered into the same group. The detailed proof is in the supplemental file.

\section{Experiments}
In this section, we first evaluate the efficiency and scalability of IMC on synthetic data. Then we show the effectiveness of the dense method PCE for IMC on a face data set. Finally we show the universality of the SDSC framework for current two-stage approaches on two handwritten digit data sets.
\subsection{Experimental Setup}
We compare the performance of current popular spectral subspace clustering methods, including OMP, SSC, SSSC, LSR, LRR, ENSC, and BDR which respectively represent $\ell_0$, $\ell_1$, $\ell_2$, mixed norm and block diagonal regularization. We use the code provided by the respective authors for computing the coefficients, where the parameters are tuned to give the best clustering accuracy. We choose BDR-Z as a representative of BDR. For those implemented in SDSC framework with different transformation functions in Table \ref{tbl:1}, the names are formed with the suffix -D1, -D2 or -D3, such as SSC-D1 indicating that the SSC algorithm is implemented with the transformation function 1). The two-stage method of directly combining IMC and spectral clustering is called as 'IMC', while the SDSC method of adding the dense stage PCE before spectral clustering is formed as 'IMC-P'. 

We evaluate the clustering accuracy (ACC) defined as:
\begin{equation}
\label{equ:14}
ACC = \frac{1}{N} \sum_{i=1}^{N} \mu\left(u_{i}, \operatorname{bestmap}\left(v_{i}\right)\right),
\end{equation}
where $u_i \in U$ and $v_i \in V$ respectively represent the output label and the ground-truth label of the $i$th data point, $\mu(x,y) = 1$ if $x = y$, and $\mu(x, y) = 0$ otherwise, and bestmap($v_i$) is the best mapping function that permutes clustering labels to match the ground-truth ones.
In addition to ACC, we also report the Normalized Mutual Information (NMI) \cite{24} and the graph connectivity (CONN), which are defined as follows:
\begin{itemize}
\item NMI is an information theoretic measure of how well the computed clusters and the true clusters predict one another, normalized by the amount of information inherent in the two clustering systems, i.e.,
\begin{equation}
\label{equ:15}
NMI(U, V)=\frac{2 \times I(U ; V)}{[H(U)+H(V)]},
\end{equation}
where $U$ and $V$ are the output labels and the ground-truth labels, $H(.)$ is entropy, $I(U;V)$ is the Mutual Information \cite{25} between $U$ and $V$. NMI is in the range $[0,1]$, and NMI = 1 stands for perfectly complete labeling.
\item
CONN evaluates the connectivity of the affinity graph. Generally, for an undirected graph with affinity matrix $W$ and degree matrix $D = \operatorname{Diag}(W\cdot\bm{1})$, where $\bm{1}$ is the vector of all ones, CONN is defined as the second smallest eigenvalue of the Laplacian $L=I-D^{-1 / 2} A D^{-1 / 2}$. CONN is in the range $[0, \frac{N-1}{N}]$ and is zero when the graph is not connected \cite{26}.
\end{itemize}

All experiments are conducted on a PC with an Intel(R) Core(TM) i7-7700 CPU at 3.60GHz, 16G RAM, running Windows 10 and MATLAB R2018a.

\subsection{Experiments on Synthetic Data}

\begin{table*}[ht]
  \centering
  \caption{ACC (\%) of different algorithms on the Extended Yale B data set. A '-' denotes time out.}
  \label{tbl:2}
  \resizebox{.84\width}!{
    \begin{tabular}{c|ccc|ccc|ccc|ccc|ccc}
    \toprule
          & \multicolumn{3}{c|}{2 subjects} & \multicolumn{3}{c|}{5 subjects} & \multicolumn{3}{c|}{10 subjects} & \multicolumn{3}{c|}{15 subjects} & \multicolumn{3}{c}{20 subjects} \\
    \hline
    methods & mean  & median & std   & mean  & median & std   & mean  & median & std   & mean  & median & std   & mean  & median & std \\
    \hline
    SSC   & 97.48 & 98.80  & 3.17  & 92.68 & 93.15 & 6.87  & 88.24 & 87.98 & 6.21  & 82.60  & 83.54 & 5.97  & 75.29 & 78.15 & 5.35 \\
    LSR   & 94.67 & 95.21 & 10.15 & 80.31 & 82.14 & 8.72  & 71.46 & 73.25 & 12.35 & 69.21 & 68.18 & 5.57  & 67.11 & 67.15 & 4.57 \\
    LRR   & 93.18 & 95.12 & 14.50  & 92.11 & 95.51 & 9.61  & 85.67 & 86.45 & 8.98  & 82.15 & 84.12 & 6.54  & 77.30  & 80.03 & 5.99 \\
    OMP   & 99.15 & 100   & 1.22  & 95.88 & 97.31 & 5.06  & 87.25 & 84.16 & 6.72  & 84.41 & 84.89 & 5.68  & 79.69 & 81.02 & 4.45 \\
    ENSC  & 91.05 & 94.22 & 15.24 & 84.98 & 84.22 & 11.24 & 77.21 & 76.24 & 10.50  & 62.97 & 65.68 & 9.24  & 63.18 & 65.48 & 5.90 \\
    BDR   & 97.31 & 97.64 & 13.54 & 87.12 & 89.35 & 11.35 & 77.23 & 79.54 & 9.54  & 69.20  & 71.24 & 5.87  & 63.63 & 66.21 & 4.54 \\
    SSSC  & 98.71 & 100   & 2.69  & 95.21 & 97.21 & 1.09  & 90.26 & 89.99 & 5.15  & 85.15 & 86.22 & 4.69  & -     & -     & - \\
    \textbf{IMC}   & \textbf{99.48} & 100   & 1.14  & 97.46 & 97.68 & 1.27  & 91.14 & 94.25 & 5.81  & 87.11 & 87.64 & 5.73  & 84.16 & 82.99 & 5.18 \\
    \textbf{IMC-P} & 99.46 & 100   & \textbf{0.81} & \textbf{97.76} & \textbf{97.76} & \textbf{1.01} & \textbf{95.38} & \textbf{95.61} & \textbf{3.97} & \textbf{91.69} & \textbf{91.11} & \textbf{3.71} & \textbf{88.13} & \textbf{86.86} & \textbf{4.12} \\
    \bottomrule
    \end{tabular}}
\end{table*}%

In this section, we evaluate the efficiency and scalability of IMC. Experiments are conducted on synthetic data.
We randomly generate 6 subspaces $\left\{S_i\right\}_{i=1}^{6}$, and the dimension of each subspace is $d_i = 6$. All the data points lie in an ambient space of dimension $D=10$. Each subspace $S_i$ contains $N_i$ data points, and $N_i$ is in the range [50, 166,667], so the total number of data points $N$ varies from 300 to 100,002. For IMC, we set the iteration number $\Gamma = 6$.
\begin{figure}[h]
\label{fig:1}
\centering
\subfigure[ACC]{
\includegraphics[width=0.47\columnwidth]{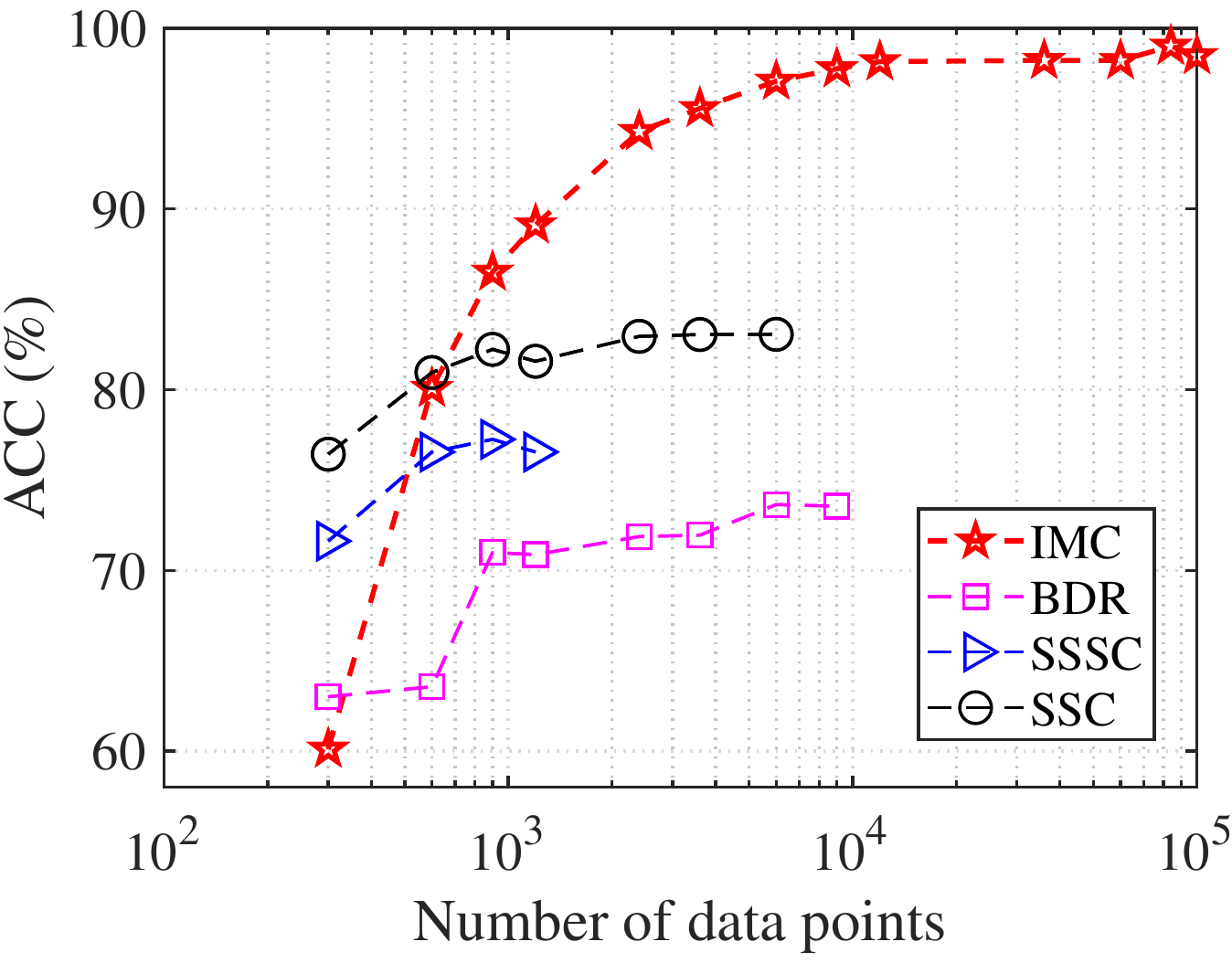}
\label{fig:1a}
}
\subfigure[NMI]{
\includegraphics[width=0.47\columnwidth]{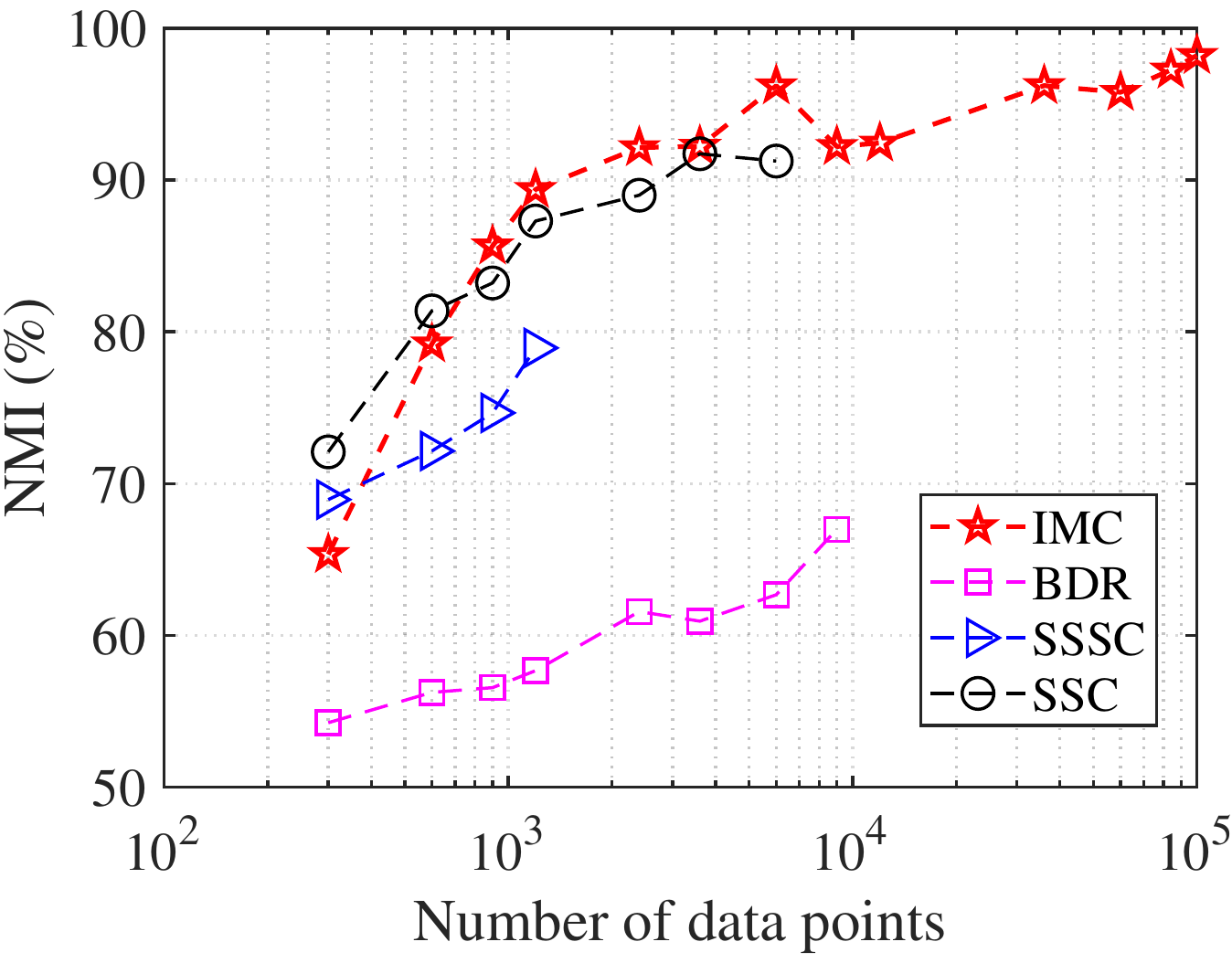}
\label{fig:1b}
}
\quad
\subfigure[CONN]{
\includegraphics[width=0.47\columnwidth]{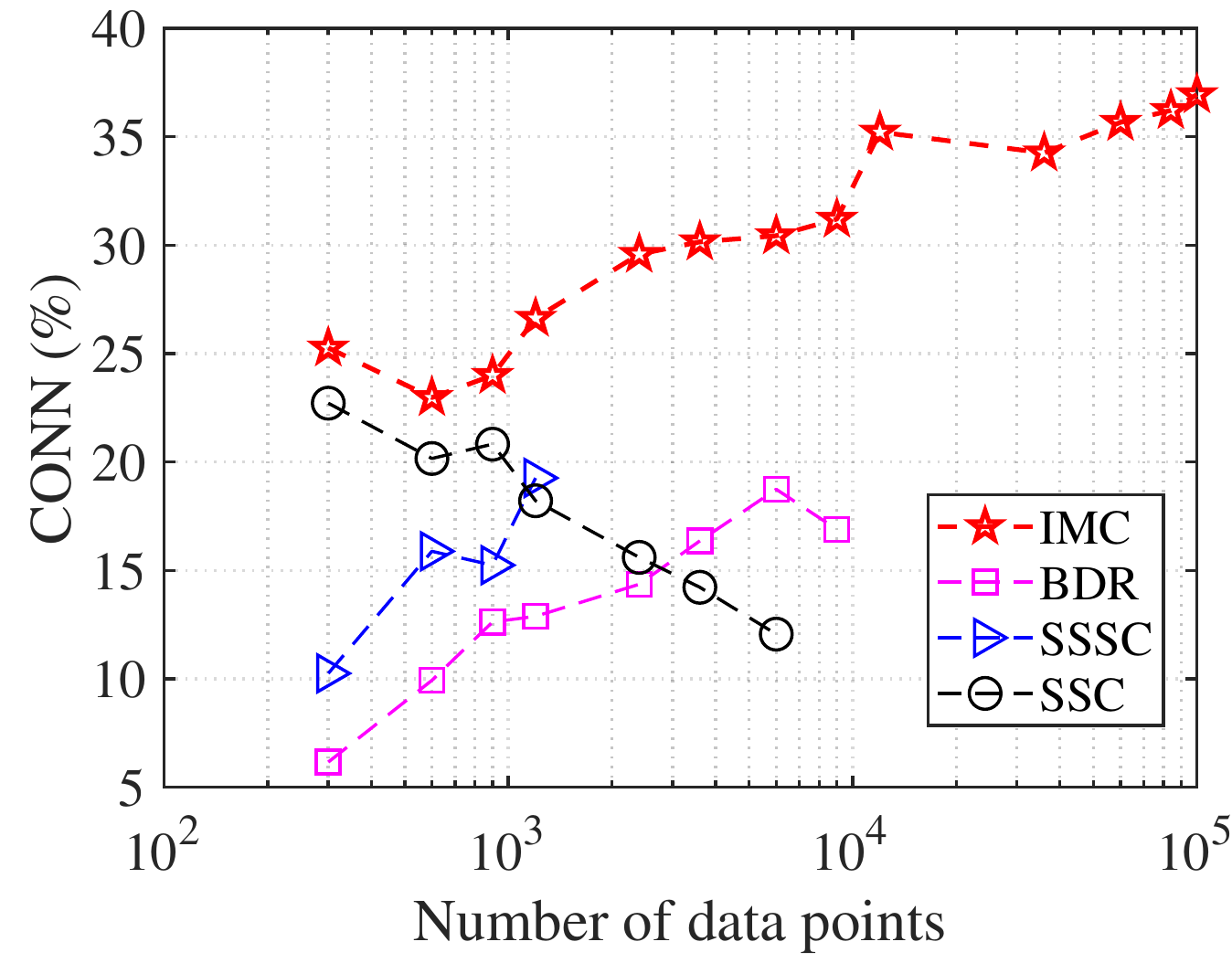}
\label{fig:1c}
}
\subfigure[Computational time]{
\includegraphics[width=0.47\columnwidth]{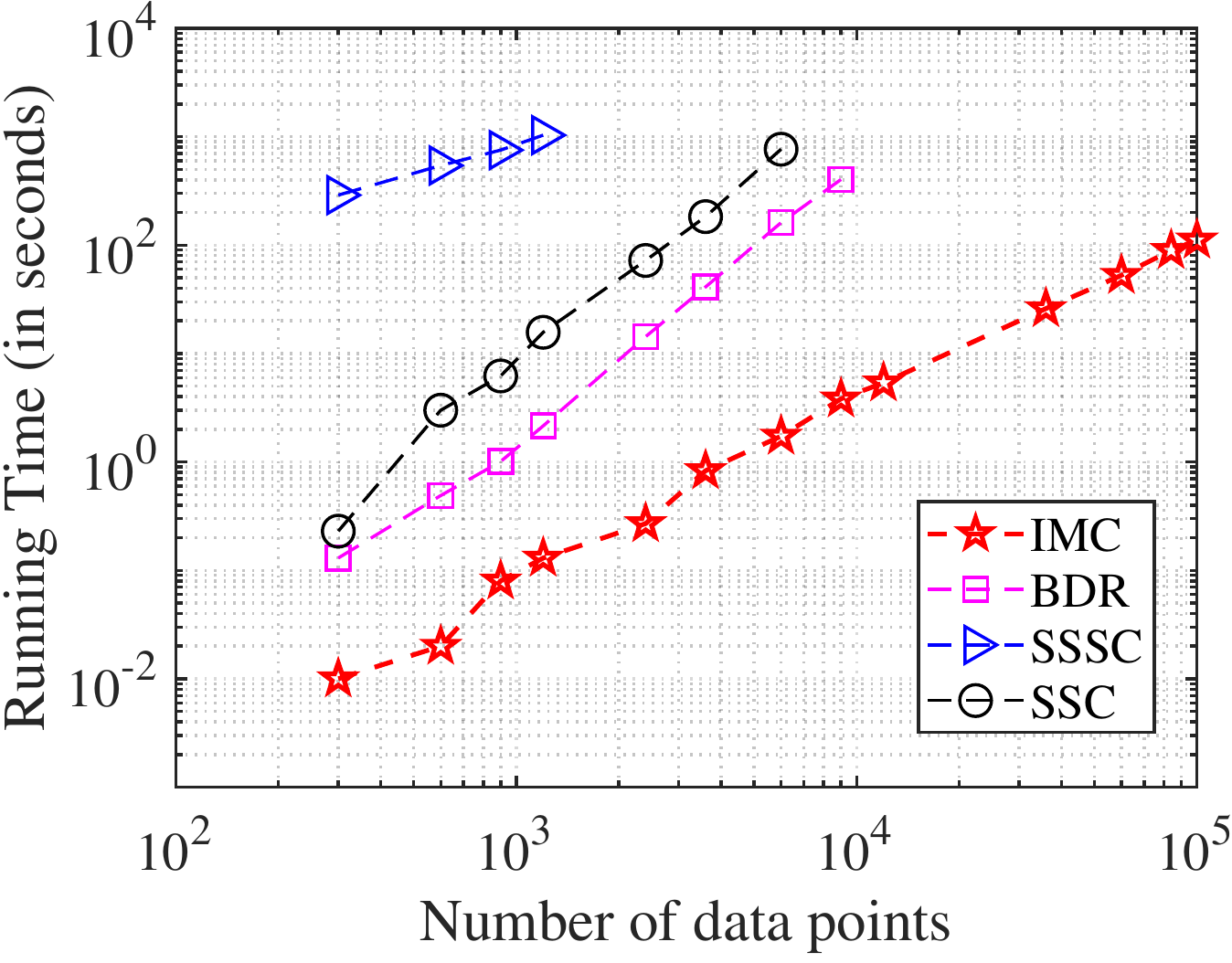}
\label{fig:1d}
}
\caption{Performance on synthetic data. For BDR, SSSC and SSC, the maximum number of points tested is respectively 9,000, 1,200 and 6,000 due to time limit. We use log scale in x-axis, and also in the y-axis of bottom right figure.}
\end{figure}

The ACC and NMI are plotted in Figure \ref{fig:1a} and \ref{fig:1b}. We first observe that IMC obtain higher accuracy and retain more information when clustering large number of data points. However, for small $N$, IMC is outperformed by SSC. This is partly because the number of connections for each data point is set as $\Gamma$, and more inter-subspace similarity are computed when $N$ is smaller. The connectivity is plotted in Figure \ref{fig:1c}, and IMC obtain the best connectivity, indicating that the points from the same subspace are well connected. The running time is plotted in \ref{fig:1d}, and it shows that IMC is significantly efficient: it is 3 orders of magnitude faster than SSC when clustering 6,000 points and 4 orders faster than SSSC when clustering 1,200 points. Actually most popular methods can only be tested on at most 10,000 points. We can conclude that as $N$ increases, the superiority of IMC in accuracy expands, and IMC generates a better sparse representation. Since IMC is significantly faster, IMC is preferable for large-scale subspace clustering problems.

\subsection{Clustering Human Face Images}

\begin{table*}[ht]
  \centering
  \caption{ACC (\%) of different algorithms on the USPS data set.}
   \label{tbl:3}
   \resizebox{.78\width}!{
\begin{tabular}{c|cccc|cccc|cccc|c}
    \toprule
    samples & SSC   & SSC-D1 & SSC-D2 & SSC-D3 & LSR   & LSR-D1 & LSR-D2 & LSR-D3 & LRR   & LRR-D1 & LRR-D2 & LRR-D3 & IMC \\
    \hline
    500   & 65.05 & 63.31 & 72.57 & \textbf{78.45} & 68.18 & 61.00 & 67.13 & \textbf{71.80} & 64.86 & 65.22 & 64.13 & \textbf{70.25} & 71.21 \\
    1000  & 60.97 & 64.75 & 69.21 & \textbf{73.13} & 71.09 & 64.15 & 70.12 & \textbf{73.79} & 61.86 & 61.42 & 60.44 & \textbf{69.02} & 69.54 \\
    2000  & 60.13 & 58.16 & 76.12 & \textbf{82.30} & 71.11 & 63.99 & 72.22 & \textbf{75.88} & 62.85 & 64.21 & 61.89 & \textbf{71.74} & 71.75 \\
    3000  & 63.54 & 56.27 & 78.73 & \textbf{83.86} & 70.94 & 62.79 & \textbf{73.46} & 72.23 & 63.72 & 65.78 & 62.94 & \textbf{67.60} & 69.88 \\
    \midrule
    samples & OMP   & OMP-D1 & OMP-D2 & OMP-D3 & ENSC  & ENSC-D1 & ENSC-D2 & ENSC-D3 & BDR   & BDR-D1 & BDR-D2 & BDR-D3 & IMC-P \\
    \hline
    500   & 61.39 & 53.25 & 67.10  & \textbf{70.57} & 60.34 & 39.28 & 66.08 & \textbf{73.87} & 66.45 & 59.12 & 69.88 & \textbf{72.16} & 72.66 \\
    1000  & 59.99 & 51.77 & 61.30  & \textbf{74.41} & 60.62 & 41.89 & 63.90  & \textbf{69.09} & 60.07 & 51.16 & 71.24 & \textbf{73.28} & 72.73 \\
    2000  & 61.62 & 49.38 & 64.89 & \textbf{73.13} & 59.21 & 44.27 & 64.97 & \textbf{71.48} & 53.54 & 44.29 & 69.00 & \textbf{72.55} & 74.21 \\
    3000  & 59.75 & 54.67 & 66.59 & \textbf{70.40} & 56.26 & 38.97 & 66.55 & \textbf{70.22} & 53.51 & 46.87 & 62.36 & \textbf{68.22} & 73.64 \\
    \bottomrule
    \end{tabular}}
\end{table*}%


\begin{table*}[ht]
  \centering
  \caption{Performance of different algorithms for clustering 5 subjects with totally 500 samples from MNIST data set.}
  \label{tbl:4}
  \resizebox{.85\width}!{
    \begin{tabular}{c|cc|cc|cc|cc|cc|cc}
    \toprule
          & SSC   & SSC-D3 & LSR   & LSR-D3 & LRR   & LRR-D3 & OMP   & OMP-D3 & ENSC  & ENSC-D3 & BDR   & BDR-D3 \\
    \hline
    ACC (\%)   & 74.97 & 92.69 & 79.48 & 88.46 & 63.10  & 77.08 & 90.25 & 93.82 & 73.51 & 89.28 & 79.95 & 93.58 \\
    NMI (\%)  & 81.39 & 85.49 & 65.42 & 77.39 & 73.25 & 80.15 & 82.89 & 84.02 & 83.02 & 86.30  & 80.09 & 88.75 \\
    CONN (\%)  & 17.31 & 79.73 & 79.05 & 89.11 & 1.07  & 65.83 & 17.01 & 74.25 & 23.01 & 79.20  & 50.81 & 85.07 \\
    \bottomrule
    \end{tabular}}
\end{table*}%
It is shown in \cite{3} that the images of a subject with a fixed pose and varying illumination approximately lie in a union of 9-dimensional subspaces. Thus subspace clustering methods can be applied on the task of face clustering. In this experiment, we evaluate the effectiveness of PCE on Extended Yale B data set. It contains 2,414 frontal face images of 38 individuals under 9 poses and 64 illumination conditions. Each cropped face image consists of 192$\times$168 pixels. We downsample the images to 48$\times$42 pixels and vectorize it to a 2,016 vectors as data points. In each experiment, we randomly pick $n \in \left\{2, 5, 10, 15, 20\right\}$ subjects and take all the images of selected subjects as data to be clustered.

\subsubsection{Performance of PCE as a Function of $\theta_1$ and $\theta_2$}
\begin{figure}[htbp]
\label{fig:2}
\centering
\subfigure[$\theta_2 = 0.6$, $\theta_1$ changes]{
\includegraphics[width=0.47\columnwidth]{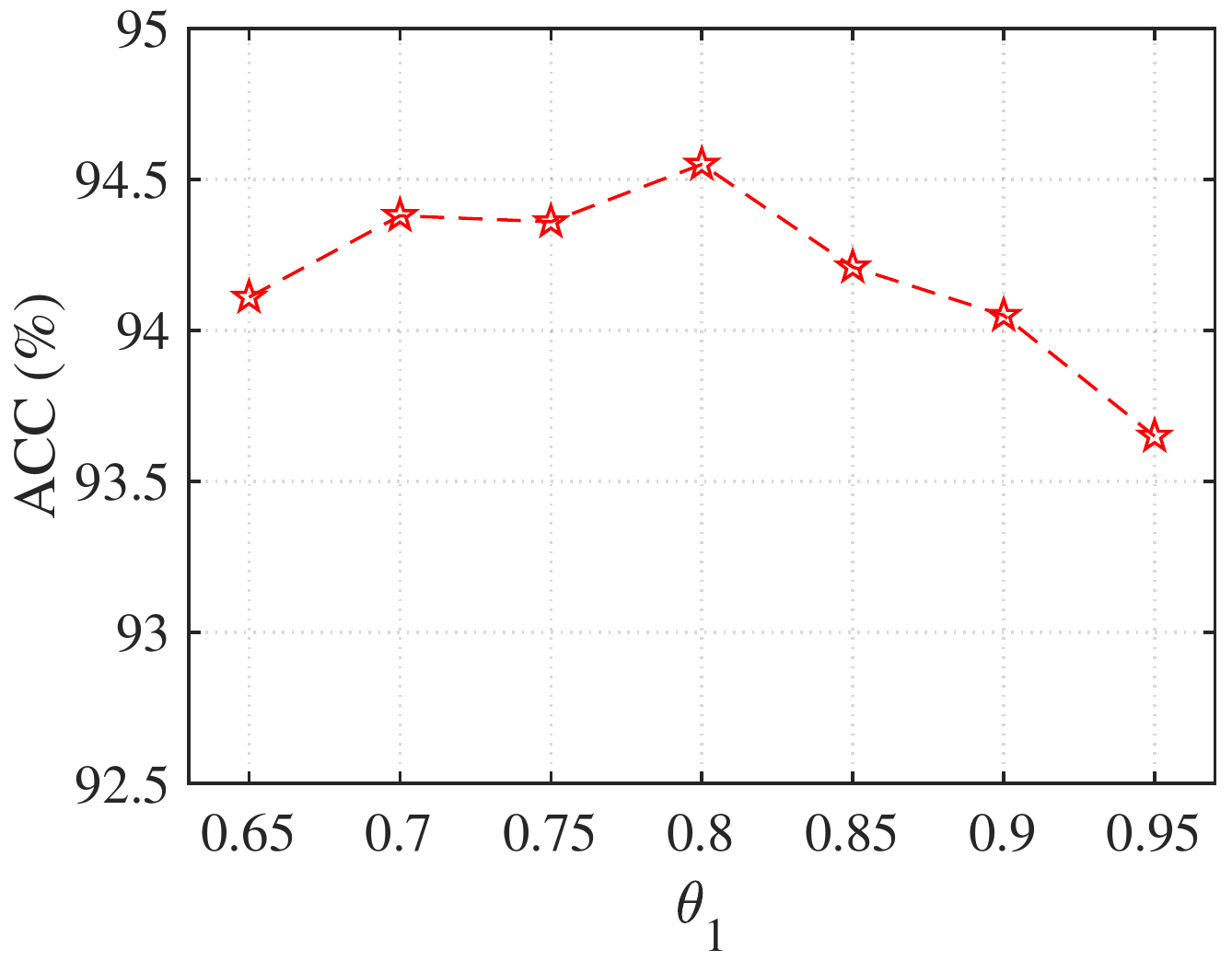}
\label{fig:2a}
}
\subfigure[$\theta_1 = 0.8$, $\theta_2$ changes]{
\includegraphics[width=0.47\columnwidth]{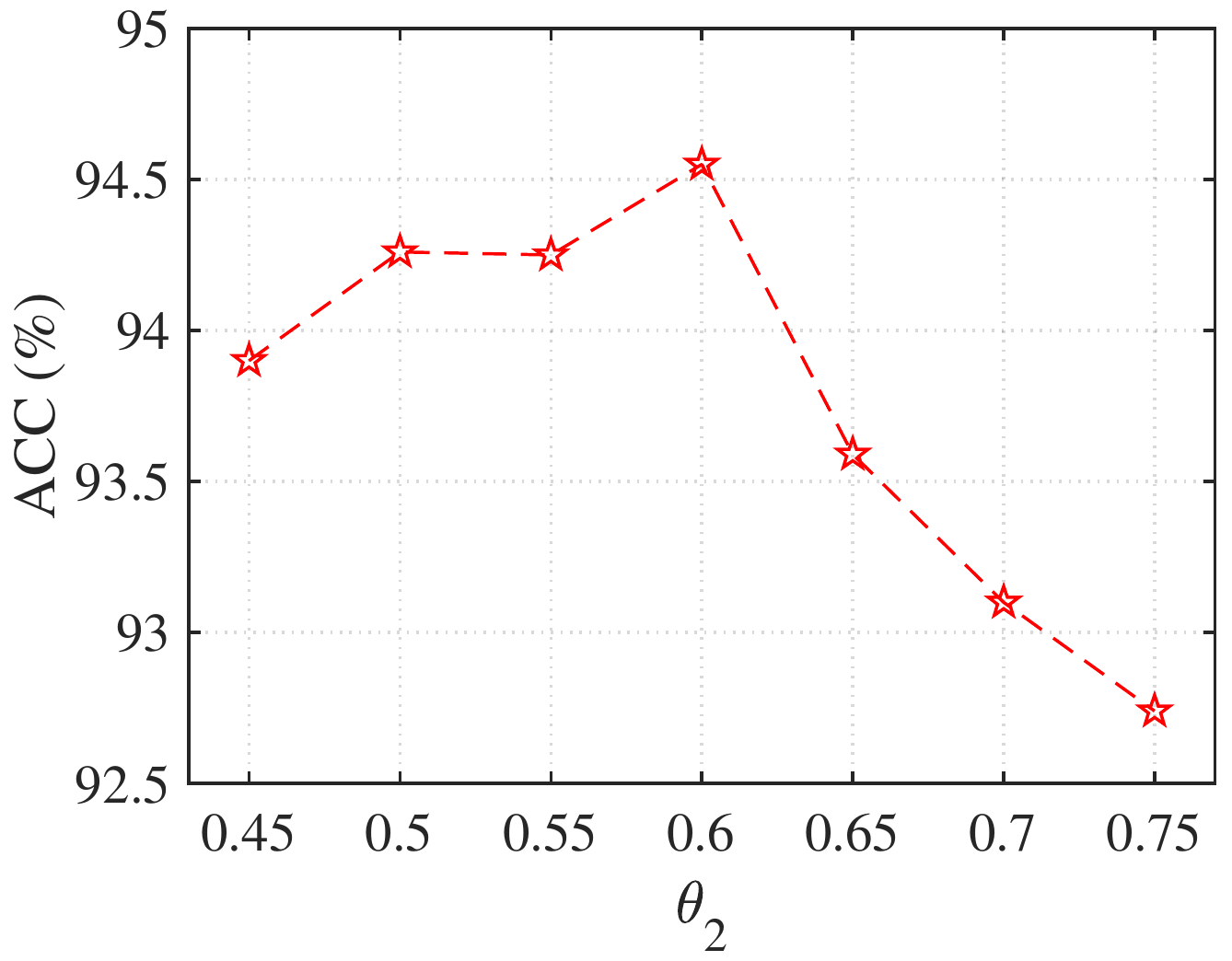}
\label{fig:2b}
}
\caption{ACC of IMC-P (IMC optimized by PCE) as a function of $\theta_1$ when fixing $\theta_2 = 0.6$ in (a) and $\theta_2$ when fixing $\theta_1 = 0.8$ in (b) for the 10 subjects clustering problem from Extended Yale B data set.}
\end{figure}

To show the effect of the parameters $\theta_1$ and $\theta_2$ on the performance of PCE, we report the average ACC on the 10 subjects problem based on two settings: (1) fix $\theta_2 = 0.6$ and choose $\theta_1 \in \left\{0.65, 0.7, 0.75, 0.8, 0.85, 0.9, 0.95\right\}$; (2) fix $\theta_1 = 0.8$ and choose $\theta_2 \in \left\{0.45, 0.5, 0.55, 0.6, 0.65, 0.7, 0.75\right\}$. The results are shown in Figure \ref{fig:2a} and \ref{fig:2b}. It can be seen that using the setting of $\theta_1 = 0.8$ and $\theta_2 = 0.6$ leads to the best accuracy, and we use this setting for following experiments.

\begin{figure}[ht]
\centering
\includegraphics[width=0.58\columnwidth]{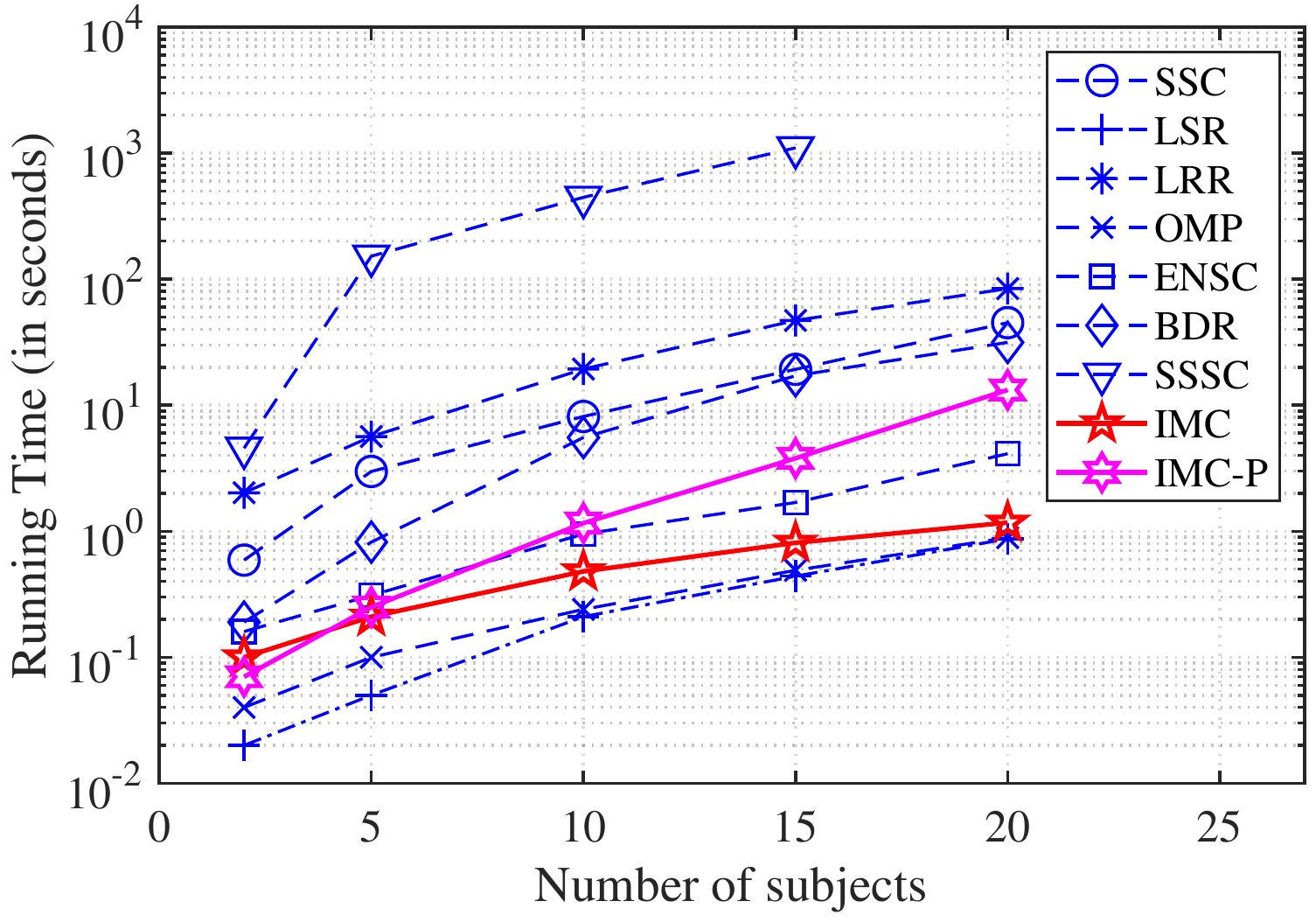}
\caption{Average computational time of different methods on the Extended Yale B data set as a function of the number of subjects. Note that we use log scale in y-axis.}\label{fig:3}
\end{figure}
\subsubsection{Performance of IMC and IMC-P Compared with Other Methods}

The clustering performance of different methods is reported in Tabel \ref{tbl:2}. It can be seen that our IMC and IMC-P outperform other methods. Generally, the clustering problem is more challenging when the number of subspaces increases. We find that when the number of subjects increases, the improvement by our methods is more significant, and the optimization of IMC by PCE expands. This experiment clearly demonstrates the effectiveness of our IMC and PCE on face clustering tasks. SSSC also performs well in some cases. However, it can be seen in Figure \ref{fig:3} that SSSC has the highest computational time, while LSR gains the lowest. Our IMC is faster than most methods, and IMC-P optimized by PCE is faster than SSSC and BDR, yet IMC-P still enjoys the highest accuracy. So our PCE helps IMC obtain a better trade-off between the performance and the time cost.

\subsubsection{Data Visualization: the Densification by PCE}
\begin{figure}[h]
\label{fig:4}
\centering
\subfigure[$W$ produced by IMC]{
\includegraphics[width=0.47\columnwidth]{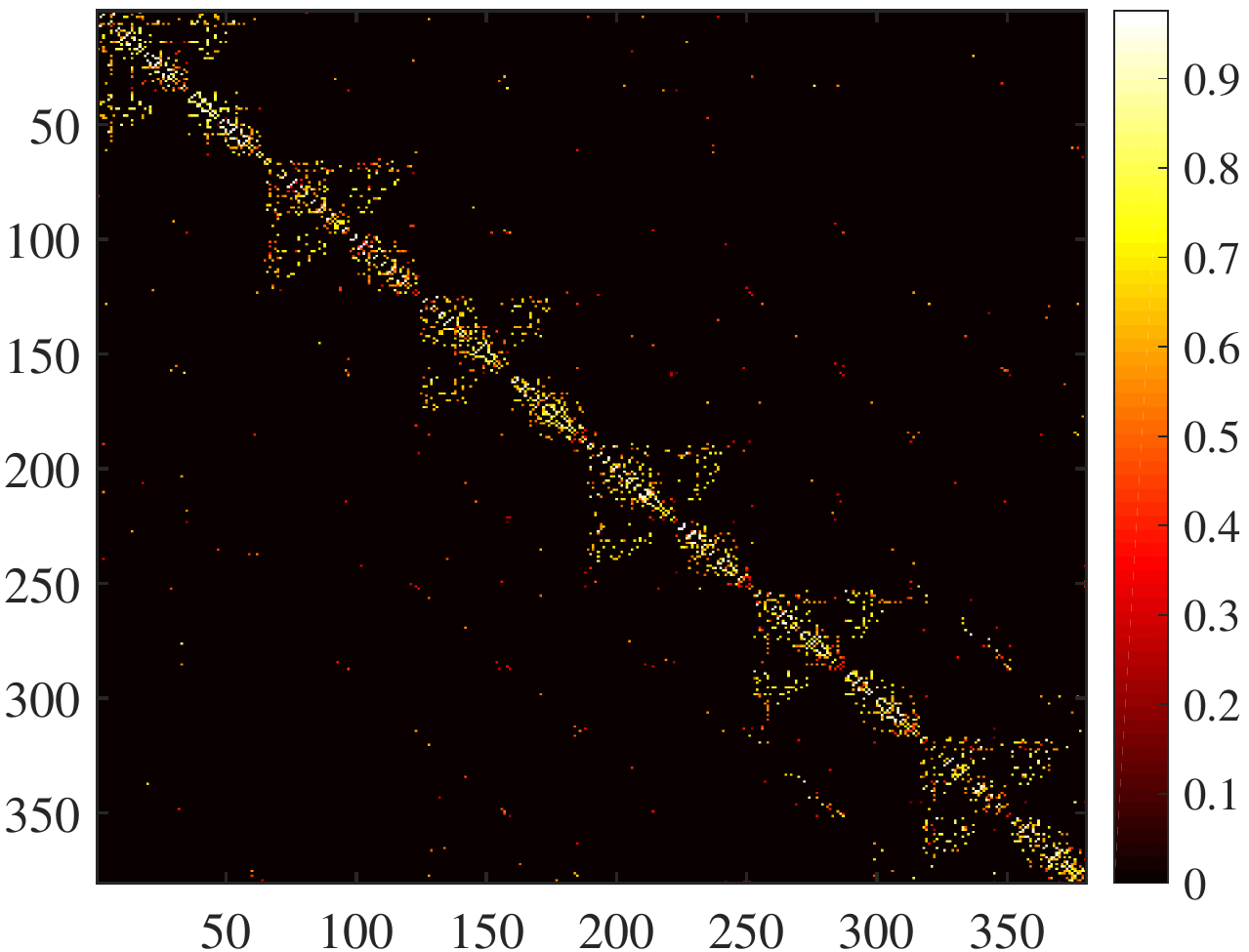}
\label{fig:4a}
}
\subfigure[$W^*$ densified by PCE]{
\includegraphics[width=0.47\columnwidth]{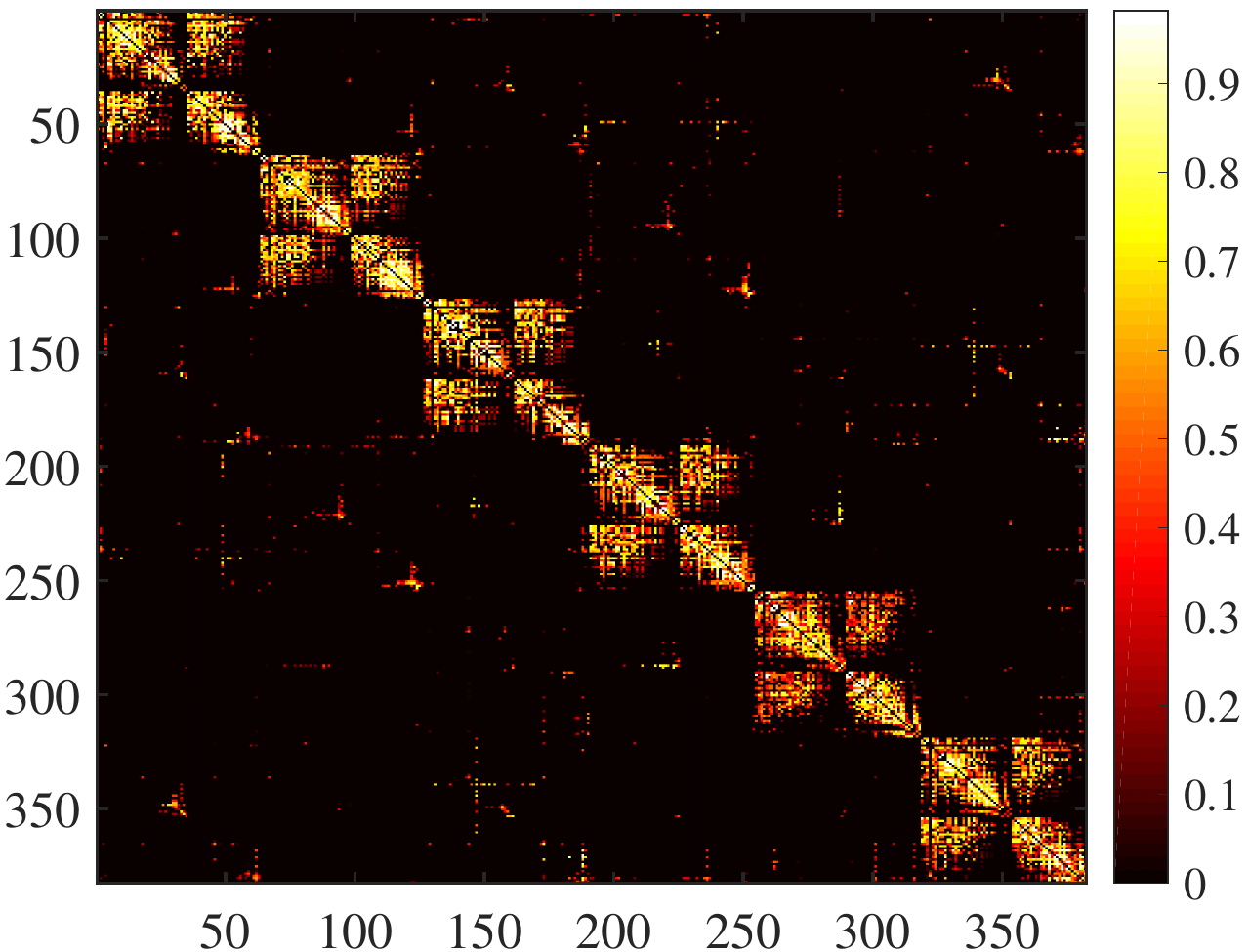}
\label{fig:4b}
}
\caption{The visualization of affinity matrix produced by IMC (a) and optimized by PCE (b) on the task of clustering 6 subjects from Extended Yale B data set.}
\end{figure}

To show the effect of using PCE to optimize the affinity matrix $W$ generated by IMC, we visualize the $W$ produced by IMC in Figure \ref{fig:4a} and $W^*$ optimized by PCE in Figure \ref{fig:4b}.
We observe that $W$ in Figure \ref{fig:4a} is a sparse matrix, and the bright dots lying in the diagonal blocks are intra-subspace similarity elements in $W$. However, it contains small amount of the intra-subspace similarity, and this may make it difficult for spectral clustering to segment the data points. After PCE, as shown in Figure \ref{fig:4b}, we gain denser intra-subspace similarity, while the number of inter-subspace elements increases little. Thus PCE is an effective optimization method for IMC to densify the intra-subspace similarity.

\subsection{Clustering Handwritten Digits}
In this part, we test the transformation functions proposed in Table \ref{tbl:1}, and verify the universality of SDSC framework. We use the USPS containing 8-bit gray scale images of handwritten digits from 0 to 9. We reshape each sample to a 200 dimension vector, and randomly pick $N_i \in \left\{100, 200, 400, 600\right\}$ samples of 5 digits, thus the total number of samples is from 500 to 3,000. Besides, we tests on MNIST with images reshaped to 500 dimension vectors.

From Table \ref{tbl:3} we can observe that the implementations suffixed with '-D3' improve the accuracy greatly in most cases, while the '-D1' methods sometimes even make the results worse. This is partly because the function 3) in Table \ref{tbl:1} could spread the value of similarity around in a more reasonable way, in which the high similarity (indicating intra-subspace similarity) are well retained and the relatively low similarity (more likely to be inter-subspace similarity) are neglected on purpose. This restricts the effect of the dense stage to the intra-subspace similarity elements in the affinity matrix. After the dense stage with function 3), the affinity matrix contain denser intra-subspace similarity, and this makes spectral clustering more likely to group points from the same subspace into the same cluster. Besides, we can notice that IMC outperforms other two-stage methods, and IMC-P improves the accuracy of it. This verifies the effectiveness of IMC and PCE on handwritten digit clustering tasks.
Moreover, Table \ref{tbl:4} reports the performance of those two-stage methods and their SDSC implementations with transformation function 3) on MNIST data set. It can be observed that the SDSC implementations obtain higher accuracy, retain more information after clustering, and get the points more connected in subspace. This demonstrates again the superiority and universality of SDSC framework. Thus SDSC is a universal framework that can be applied on current two-stage methods to get better performance.
\section{Conclusion and Future Work}
This paper studies the subspace clustering problem which aims to clustering the high-dimensional data points into low dimension according to the self-expressiveness model. We first propose a new faster sparse method Iterative Maximum Correlation (IMC) to draw coefficients from data, then apply a dense method Piecewise Correlation Estimation (PCE) to optimize the affinity matrix. IMC adopts the Pearson correlation coefficient as both a measure to select points and the value of similarity. PCE optimizes the similarity between two data points via the intermediate data points. Besides, we extend our work to be a SDSC framework for current two-stage subspace clustering approaches. To the best of our knowledge, we are the first one to densify the affinity matrix before spectral clustering. We show the efficiency and scalability of IMC when handle 100,000 data points, whereas most approaches have only be tested less than 10,000 points. We also show the effectiveness of PCE to densify the intra-subspace similarity. We also make analysis of our SDSC framework. We conduct series of experiments to demonstrate the effectiveness of our methods. We note that our universal dense method in Algorithm \ref{alg:4} optimize the similarity based on graph analysis of distance, and other styles of dense methods are left for future research.

\bibliographystyle{aaai}
\bibliography{bibfile}
\end{document}